\documentclass[letterpaper]{paper}
\usepackage{AAAI}
\usepackage{times}
\usepackage{helvet}
\usepackage{epsfig}
\usepackage{courier}
\usepackage[noeepic]{qtree}
\usepackage{enumitem}
\usepackage{gb4e}
\usepackage{graphicx}
\usepackage[all]{xy}

\begin{document}
\title{Dependency Parsing with Dynamic Bayesian Network}
%\\ \hspace{30pt}\\\small Content areas: Bayesian Networks, NLP, Cognitive Modeling\\Note: Closely related paper is under review for CogSci 2005 \\ ID 579}
\author{Virginia Savova \\
Department of Cognitive Science \\ Johns Hopkins University \\3400 N. Charles St.\\Baltimore MD 21218
\And
Leonid Peshkin \\ Department of Systems Biology \\Harvard Medical School\\200 Longwood Ave\\ Boston MA 02115}

\maketitle
\begin{abstract}
\begin{quote}
Exact parsing with finite state automata is deemed inappropriate because of the unbounded non-locality languages overwhelmingly exhibit.  We propose a way to structure the parsing task in order to make it amenable to local classification methods. This allows us to build a Dynamic Bayesian Network which uncovers the syntactic dependency structure of English sentences. Experiments with the Wall Street Journal demonstrate that the model successfully learns from labeled data. \end{quote}
\end{abstract}

\section{Introduction} 

Bayesian graphical models have become an important explanatory strategy in cognitive science (\cite{Knill96},\cite{Kording04}, \cite{Stocker05}). Recent work  strongly supports their biological plausibility in general and that of dynamic Bayesian models in particular \cite{Rao05}. Dynamic models are geared towards prediction and classification  of sequences. As such, they are naturally suitable for language modeling and have already been aplied to tasks like speech recognition \cite{Bilmes03} and part-of-speech tagging \cite{Peshkin03}. However, grammar learning and parsing with such models generally appears out-of-reach, because of their Markovian character. 

Markov models restrict possible dependencies to a bounded, local context. At one extreme, the context is confined to the symbol occupying the current position in the sequence (order-0 or unigram models). In more relaxed versions, context may include a fixed number of positions before the current symbol (k-order), typically no more than three (trigram models). The restricted space of possible dependencies allows transition probabilities to be infered from the data and stored in a look-up table with relatively little technical sophistication. 

Not surprisingly however, the restricted space of representable dependencies is also the main disadvantage of Markov models in syntax-related tasks like parsing. Syntactic dependencies in natural language are unboundedly non-local, in the sense that no fixed amount of context is guaranteed to contain the members of a given constituent. For example, consider the sentences in examples (\ref{exa:nl1} - \ref{exa:nl3}). In the first sentence, the subject \emph{king} and verb \emph{bought} are adjacent to one another. Thus, the dependecy between them would be captured by a bigram (order-1) model. However, the same model would be unable to represent the dependendency in the second example, because the subject and verb are separated by two words. To capture this dependency, we need a 3rd-order Markov model. Similarly, the 3rd-order model would prove inadequate for the third example, where the subject and verb are separated by four words.

\begin{exe}
\ex  The king bought a camel.
\label{exa:nl1}
\end{exe}

\begin{exe}
\ex  The king of Prussia bought a camel.
\label{exa:nl2}
\end{exe}

\begin{exe}
\ex  The king of some strange country bought a camel.
\label{exa:nl3}
\end{exe}

Our solution to this problem relies on representing sentences with non-local dependencies like (\ref{exa:nl2}, \ref{exa:nl3}) as derived from their local dependency variants, akin to (\ref{exa:nl1}). This intuition is based on the formal notion that a string with non-local dependency is obtained from a dependency tree via a recursive linearization procedure. The string obtained at each step of the linearization procedure contains new local dependencies, which push apart local dependencies from previous levels. This way of conceptualizing the linearization of syntactic structure allows us to use a Dynamic Bayesian Network despite its Markov properties. We construct a DBN parser which decides only on local attachments. We then call the parser recursively to uncover the underlying dependency tree. Our results show that the model captures grammatical knowledge for all levels of the derivation. The biological plausibility and remarkable compactness of learned representation may suggest that parsing in the brain is accomplished in a similar manner. 

\section{Dependency grammar} 

\begin{figure}[h]
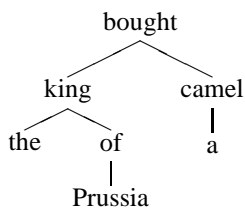

\qtreecenterfalse
\Tree [.{bought} [.{king} the [.of Prussia ]] [.{camel} a ] ]
\par
\caption{Dependency structure of example (\ref{exa:nl2})}
\label{fig:tree}
\end{figure}

Tree-based linguistic representations of natural language syntax treat non-local dependencies as local in the two-dimensional tree structure, of which the string is a one dimensional projection. The dependency grammar representation of (\ref{exa:nl1}) captures the dependency between the subject, the object and the verb, and the dependency between the determiners and their respective nouns (Figure \ref{fig:tree}). 
%Arguably, the most theory-neutral tree representation is 

More formally, a dependency grammar consists of a lexicon of terminal symbols (words), and an inventory of dependency relations specifying inter-lexical requirements. A string is generated by a dependency grammar if and only if:  
\begin{itemize}
	\item \vspace{-.2cm} Every word but one (ROOT) is dependent on another word. 
	\item \vspace{-.2cm} No word is dependent on itself either directly or indirectly. 
		\item \vspace{-.2cm} No word is dependent on more than one word. 
		\item \vspace{-.2cm} Dependencies do not cross.
\end{itemize}
In a dependency tree, each word is the mother of its dependents, otherwise known as their {\sc  head}. To linearize the dependency tree in Figure \ref{fig:tree} into a string, we introduce the dependents recursively next to their heads: \\
Step I: bought \\  
Step II: king bought camel \\ 
Step III: The king of bought a camel\\
Step IV: The king of Prussia bought a camel. 

\section{Recursive parsing as local classification}

Parsing in the dependency grammar framework is the task of uncovering the dependency tree given the sentence. Suppose that instead of searching for a complete parse given a complete sentence, we restricted our task to compressing the string up the linearization path. Note that linearization is essentially dependency parsing in reverse. In other words, we can uncover the dependency structure by labeling the \textbf{local} head-dependent relationships at the bottom linearization level (i.e. the sentence) and erasing from the string the words whose heads are already found. We recursively process the output until the root level. Thus, if as a first step in parsing (\ref{exa:nl2}), we pick the head of \emph{Prussia} to be the preposition \emph{of}, we can compress the string to a form virtually equivalent to linearization Step III. Picking \emph{king} as the head of the preposition leads us to compress the string further, to the equivalent of step II. To compress the string, we must simply identify which words in the string occupy a position adjacent to their heads.

The attractive feature of this representation is that the parsing decisions taken at each step are local. Hence, parsing can be converted into a local classification task. The task is to chose the best sequence of labels denoting local dependency relationships (links). At each position, we choose between setting the link to {\sc  left}, {\sc  right}, or {\sc  none}, where {\sc  left}/{\sc  right} means the word is dependent on its left/right neighbor. {\sc  none} means the search for this word's head should be postponed until later stages of compression. The output of the classifier is a labeled string, which can be compressed by removing linked dependents. It is fed through recursively, until the string is compressed to the ROOT.

\section{The Dynamic Bayesian Network classifier} 
 
The first step towards building the classifier is coming up with a feature representation. We will briefly motivate the choice of feature set with linguistic arguments. It is easy to determine that the linking pattern of a word depends on its part of speech (PoS) and the part of speech of its neighbor. For example, English determiners only link to the right, and adverbs link almost exclusively to verbs. However, the parts of speech alone are not sufficient to determine linking behavior. In some cases, the identity of the adjacent word is required -  \emph{bought} accepts links from nouns to the right, while \emph{slept} does not. 

Another decisive factor is how many dependents the current word has acquired so far. Since once the current word is linked it will become unavailable as a future linking target to other words, it is important to acertain that its valency has already been satisfied. Valency refers to the \textbf{minimal} number of dependents a word actively seeks to license. In English and other SVO languages, the word has particular requirements with respect to the number of left and right dependents. Thus, in our feature representation, valency is indirectly captured by two variables, which reflect the number of dependents which had already been linked to the current word from either side - {\sc left} and {\sc right composite (comp)}. The {\sc comp} variables affect not only the linking behavior of the current token, but that of its neighbor as well. If the word has already received many dependents from one side, the probability of accepting yet another one becomes smaller, since its valency is already satisfied.

 Finally, the current label depends on the labels of its neighbor, because if the previous label is {\sc  right}, then the current label cannot be {\sc  left}, and if the next label is {\sc  left}, the current label cannot be {\sc  right}. Thus, our full feature representation consists of the word and its PoS tag, the words and PoS tags of its neighbor, the two valencies of the current word, the right valency of its left neighbor and the left valency of its right neighbor, as well as the neighboring links. 
 
The Word and Next Word feature vocabulary contain the $2500$ most frequent words in the data. An additional value was allocated for all remaining out-of-vocabulary words. The PoS, and Next PoS vocabulary contain $36$ of the original $45$ Penn Treebank Tagset, after all punctuation $PoS$ tags were removed. The left and right {\sc comp} features had tree values: {\sc none}, {\sc one} and {\sc many}.
 
This feature representation is used as the basis of the Dynamic Bayesian Network ({\sc dbn}). After we briefly introduce the essential aspects of {\sc dbn}s, we wil expand on the structure of the network for parsing. For more information on the general class of models, we refer the reader to a recent dissertation \cite{Murphy02} for an excellent survey. 

\subsection{General notes on DBNs}

A {\sc dbn} is a Bayesian network unwrapped in ``time'' (i.e. over a sequence), such that it can represent dependencies between variables at adjacent position. More formally, a {\sc dbn} consists of two models $B^0$ and $B^+$, where $B^0$ defines the initial distribution over the variables at position $0$, by specifying:

\begin{itemize}
\item set of variables $X_{1} , \ldots, X_{n}$;
\item directed acyclic graph over the variables;
\item for each variable $X_{i}$ a table specifying the conditional\\ probability of $X_{i}$ given its parents in the graph $Pr(X_{i}|Par\{X_i\})$.
\end{itemize}
The joint probability distribution over the initial state is: 
	\[\Pr(X_1,...,X_n)=\prod_1^n\Pr(X_i|Par\{X_i\}).\]
The transition model $B^+$ specifies the conditional probability distribution ({\sc cpd}) over the state at time $t$ given the state at time $t\!-\!1$. $B^+$ consists of: 
\begin {itemize}
\item directed acyclic graph over the variables $X_1,\ldots,X_n$ and their predecessors $X_1^-,\ldots,X_n^-$ --- roots of this graph;
\item conditional probability tables $\Pr(X_i|Par\{X_i\})$ for all $X_i$~(but not $X_i^-$).
\end {itemize}
The transition probability distribution is: 
\[\Pr(X_1,...,X_n\Big|X_1^-,...,X_n^{-})=\prod_1^n\Pr(X_i|Par\{X_i\}).\]
Together, $B^0$ and $B^+$ define a probability distribution over the realizations of a system through time, which justifies calling these {\sc bn}s ``dynamic''. In our setting, the word's index in a sentence corresponds to time, while realizations of a system correspond to correctly tagged English sentences. Probabilistic reasoning about such system constitutes inference. 

Standard inference algorithms for {\sc dbn}s are similar to those for {\sc hmm}s. Note that, while the kind of {\sc dbn} we consider could be converted into an equivalent {\sc hmm}, that would render the inference intractable due to a huge resulting state space. In a {\sc dbn}, some of the variables will typically be observed, while others will be hidden. The typical inference task is to determine the probability distribution over the states of a hidden variable over time, given time series data of the observed variables. This is usually accomplished using the forward-backward algorithm. Alternatively, we might obtain the most likely sequence of hidden variables using the Viterbi algorithm. These two kinds of inference yield resulting {\sc link} tags. 

Learning the parameters of a {\sc dbn} from data is generally accomplished using the EM algorithm. However, in our model, learning is equivalent to collecting statistics over cooccurrences of feature values and link labels. This is implemented in {\sc gawk} scripts and takes minutes on a large corpus. While in large {\sc dbn}s, exact inference algorithms are intractable, and are replaced by a variety of approximate methods, the number of hidden state variables in our model is small enough to allow exact algorithms to work. For the inference we use the standard algorithms, as implemented in a recently released toolkit \cite{BilmesSoft}.

\subsection{Structure of the DBN parser}

\begin{figure}[h]
\begin{center}
\vspace{0.5cm}
\includegraphics[width=7cm]{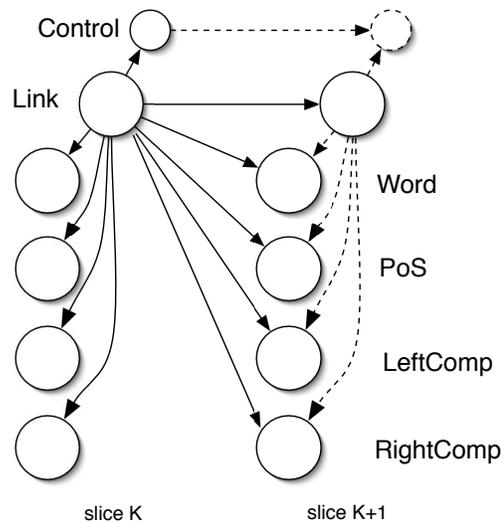}
\end{center}
\caption{The parsing DBN.} 
\label{fig:DBN}
\end{figure}

Each slice of our DBN parser is a representation of the joint probability distribution of {\sc word, pos, left/right comp}, and the hidden variable {\sc link} Figure~\ref{fig:DBN}. In our model, the link determines the value of all variables and they are independent of one another. Of course, this is not truly the case, but among those variables {\sc link} is the only unobserved, hence modeling all other dependencies is inconsequential. In addition  to the intra-slice dependencies, we model dependencies between the current, previous and next position.  The {\sc link} variable infulences all aforementioned variables in neighboring slices. Finally, we introduce a {\sc control} variable which deterministically ensures that at least one link in the sequence will be set to something other than {\sc none}. This forces the parser to trully compress the string at each recursive parsing step.   

\section{Experiments and results} 

For the results presented here we used the WSJ10 corpus \cite{Klein04}. It is a subset of the WSJ Penn Treebank (\cite{Marcus93}), consisting of all sentences shorter than eleven words with punctuation removed \footnote{the dot in our figures stands for an abstract ROOT symbol}. The dependency annotation was obtained through automatic conversion of the original treebank annotation. The relatively short sentences make this corpus a good approximation to casual speech and limit the effects of misattachments due to the conversion. 
\subsection{Encoding}

\begin{figure}[h]
\begin{flushleft}
\vspace{0.1cm}
\includegraphics[width=9cm]{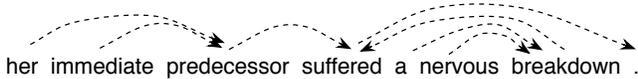}
\end{flushleft}
\caption{Dependency structure.} 
\label{fig:enc}
\end{figure}

\begin{figure}[h]
\begin{flushleft}
\vspace{0.1cm}
\includegraphics[width=9cm]{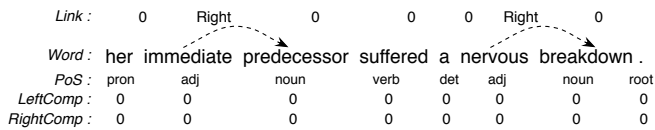}
\end{flushleft}
\caption{First layer representation.} 
\label{fig:encfirst}
\end{figure}

\begin{figure}[h]
\begin{center}
\vspace{0.1cm}
\includegraphics[width=7cm]{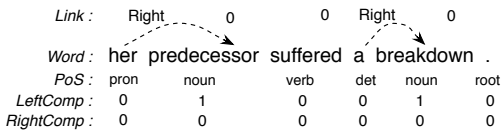}
\end{center}
\caption{Second layer representation.} 
\label{fig:enc3}
\end{figure}

\begin{figure}[h]
\begin{center}
\vspace{0.1cm}
\includegraphics[width=6cm]{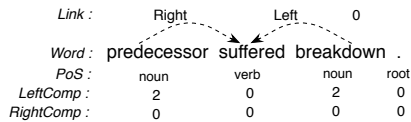}
\end{center}
\caption{Third layer representation.} 
\label{fig:enc4}
\end{figure}

\begin{figure}[h]
\begin{center}
\vspace{0.1cm}
\includegraphics[width=4cm]{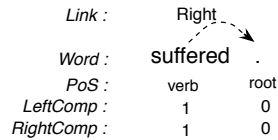}
\end{center}
\caption{Top layer representation.} 
\label{fig:enclast}
\end{figure}

The corpus was encoded in our feature representation as follows. For each sentence, a number of feature files were produced containing the feature representation of the sentence at each linearization level. The encoding of an actual sentence-structure pair from our corpus (Figure \ref{fig:enc}), is illustrated in Figures \ref{fig:encfirst} to \ref{fig:enclast}.\\ 

At the lowest level, no word has any discovered dependents, hence the {\sc comp} values are zero everywhere. All links of words whose heads are not adjacent are labeled {\sc none (0)}.\\
At the next level, words whose labels were {\sc left} or {\sc right} are removed from the structure and the {\sc comp} counters for their head are incremented.\\
 \\
The same procedure produces the subsequent levels (Figures \ref{fig:enc4}, \ref{fig:enclast})

\subsection{Testing}
The corpus was split randomly 9:1 into a training and testing section. In training mode, the DBN was given all levels with the correct labels. It was trained directly on the annotations, with no additional smoothing. The result achieved was $79\%$ correct link attachment for directed dependencies, and $82\%$ for undirected. We compare the results to two baselines given for this corpus by \cite{Klein04}, Table~\ref{tab:comp}.  

\begin{table}[!ht]
\begin{center} 
\caption{DBN results against baseline.} 
\label{tab:comp} 
\vskip 0.12in
\begin{tabular}{lll} 
\hline
Model    & Accuracy & \\
\hline
         & Dir & Undir\\
DBN       &   79 & 82 \\
Random  &   30 & 46 \\
Adjacent heuristic  &  34 & 57 \\
\hline
\end{tabular} 
\end{center} 
\end{table} 

More detailed results for our model are shown in Table~\ref{tab:allres} .

\begin{table}[!ht]
\begin{center} 
\caption{Detailed results for the DBN.} 
\label{tab:allres} 
\vskip 0.12in
\begin{tabular}{ll} 
\hline
Measure  & Accuracy  \\
\hline
Root dependency         & 83\\
Non-root dependency      & 78 \\
Out-of-Vocabulary  & 75 \\
Sentence & 36\\
\hline
\end{tabular} 
\end{center} 
\end{table} 
 
The results unequivocally surpass the random baseline and the best available heuristic, which amounts to linking every word to its right neighbor. This suggests our model has learned at least some of the non-trivial dependencies which govern the choice of link structure. The minimal difference between the vocabulary and out-of-vocabulary scores imply that the network can recover the syntactic properties of an unknown word in context. The fact that the root accuracy is higher than the non-root accuracy allows us to conclude that the network correctly learns to postpone decisions about the root word in all cases, and about its dependent in most cases.

\section{Discussion}

Our results show that combining a DBN model with recursive application is a reasonable parsing strategy.  This opens the door to the hypothesis that Bayesian inference is a possible mechanism for parsing in the brain, despite the Markovian properties of the corresponding dynamic models. The high ROOT accuracy suggests that the model has captured some fundamental principles defining the local dependency structure at all levels of the derivation. We take this result as evidence that graphical models with Markov properties are capable of handling unbounded non-local dependencies through recursive calls on their own output. The implication of this finding transcend Bayesian graphical models and speak to the general issue of how relevant other biologically plausible Markov models can be to language processing and learning. For example, Elman networks have been criticized for their a priori limitation in handling unbounded dependencies \cite{Frank}. It is possible that such type of models may be adapted to discover locality in the hierarchical structure through recursive application.

One exciting implication of this hypothesis is the domain-generality of Bayesian inference and learning mechanisms. Previous work has proposed that these mechanisms are involved in visual perception \cite{Knill96}, \cite{Kersten03}, motor control \cite{Kording04}, and attention modulation \cite{Yu05}. \cite{Kersten03} proposes Bayesian graphical model of object detection which rely on estimating hidden variables such as relative depth and 3-D structure from observables they influence -shadow displacement, 2-D projection. \cite{Kording04} suggests that subjects ina sensory-motor experiment internally represent both the statistical distribution of the task and their sensory uncertainty, combining them in a manner consistent with a performance-optimizing bayesian process. In our work, the hidden links are estimated from  observable word and PoS, along with a prior label distibution. 

The parallelism in the proposed cognitive strategies for all these different modalities may shed light on the issue whether and how modular the language faculty is. The modularity hypothesis states that the cognitive mechanisms underlying linguistic competence are specific to language. If Bayesian inference proves to be a plausible uniting principle behind visual, motor and linguistic abilities, this hypothesis is seriously undermined. At the same time, it is important to note that the generality of the mechanism does not necessarily negate the modularity of language completely. The feature representation which our model used already encodes language-specific knowledge. Further research is needed to determine whether the feature representation and the structure of the network can be induced through structure learning algorithms.

Our approach is particularly appealing in light of recent work suggesting that Bayesian type inference is biologically plausible. \cite{Rao05} shows that recurrent networks of noisy integrate-and-fire neurons can perform approximate Bayesian inference for dynamic and hierarchical graphical models. According to him, the membrane potential dynamics of neurons corresponds to approximate belief propagation in the log domain, and the spiking probability of a neuron approximates the posterior probability of the preferred state encoded by the neuron, given past inputs. This seems to suggest that our parsing model can be implemented in a neural circuit. Furthermore, since the same DBN is used to uncover local dependencies throughout all levels of the derivation, such implementation would address Humboldt's characterization of language as a system that makes ``infinite use of finite means'' at the neurophysiological level. The same neural aparatus could be used to recursively uncover the dependency structure of a sentence level by level. 

Another implication of our work is that the nature of the processing architecture may constrain the kind of grammar human languages permit. If indeeed parsing is accomplished through recursive processing of the output of previous stages, some types of long-distance depndencies would be impossible to detect. In particular, if the material intervening between a head-dependent pair (H, D) is not a constituent whose own head depend on either H or D, our model would not be able to uncover it because H and D will not be adjacent at any point in the derivation. In other words, this parser is incapable of handling strictly context-sensitive languages. to the extent that such dependencies exist, they are fairly limited \cite{Shieber85}. Such cases will need to be resolved through some reordering in pre-processing, possibly based on case marking. 
   
\section{Future work}

One deficiency of our model is that decisions at lower levels cannot be reversed in the  interest of more optimal choices at higher levels. There are however important reasons why this might be necessary. For example, a prepositional phrase subcategorized for by the verb may be mistakenly attached to a preceding noun phrase, leaving the verb with a missing dependent (\ref{exa:put}) 

\begin{exe}
\ex  The king put *[the camel in the trunk].
\label{exa:put}
\end{exe}
In the future, we hope to address this problem through a form of beam search - retaining the k-best parses at each level and choosing among them based on what happens at the next level.

Another important issue that we need to address is the total loss of information about the dependents that have been linked to a word at previous levels. Some well-known cases pose a problem for this aspect of our model. For example, the sentences in (\ref{exa:tel}) and (\ref{exa:hump}) are structurally distinct solely becase the complement of the prepositional phrase in the second sentence is an instrument appropriate for seeing. 

\begin{exe}
\ex  The king saw [the camel with two humps] .
\label{exa:hump}
\end{exe} 

\begin{exe}
\ex  The king saw *[the camel with a telescope].
\label{exa:tel}
\end{exe}

In our current model, once the complement is linked to the preposition, the two sentence will become identical, and one of them will be assigned the wrong structure. This concern can be addressed through introducing new variables, which keep track not only of the number of linked dependents but of their semantic category (e.g. instrument, animate etc.)

A natural way to extend our model in a different direction is to combine it with the Bayesian PoS tagger developed in \cite{Peshkin03}. Allowing the model to infer PoS tags and structure simultaneously will be a significantly better approximation to the parsing task humans are faced with. Last but not least, we would like to implement semisupervised learning. One way to do this would involve starting off with a small labeled set of sentences at all parsing depths, followed by presenting unparsed whole sentences. The parses suggested by the model would in their turn be used for learning in a bootstrap fashion. 

\section{Conclusion}

In our closing remarks, we would like to emphasize several aspects of our parsing model which make it interesting from the perspective of cognitive science and brain-inspired artificial intelligence. First, it belongs to a class of models which have been used recently to capture cognitive mechanisms in non-linguistic domains. Second, it naturally utilizes the overwhelming ``disguised locality'' of natural language syntax - in other words, it benefits from the fact that string-non-local dependencies are tree-local. Third, it is biologically plausible because it has been shown to be implementable in a neural circuit. And finally, it takes seriously the question how the finite amount of brain hardware is capable of encoding structures of unbounded depth. While there is much room for improvement, we believe these qualities make it an important step on the difficult road toward understanding how the mind emerges from the brain. 
\bibliography{AAAI05}
\bibliographystyle{apalike}
\end{document}